\documentclass[review]{elsarticle}
\usepackage{subfigure}
\usepackage{lineno}
\usepackage{hyperref}
\usepackage{amsfonts,amsmath,bm}
\usepackage{mathrsfs}
\usepackage{algorithm}
\usepackage{algorithmic}
\usepackage{multirow}
\usepackage{amsmath} 
\usepackage{amsthm}
\usepackage{indentfirst}
\usepackage[normalem]{ulem}
\useunder{\uline}{\ul}{}
\newtheorem{theorem}{Theorem}[section]

\newtheorem{lemma}{Lemma}[section]
\journal{Journal of \LaTeX\ Templates}

\begin{document}

\begin{frontmatter}

\title{Unsupervised Domain Adaptation with Progressive Adaptation of Subspaces}


\author[mymainaddress,mysecondaryaddress]{Weikai Li}

\author[mymainaddress,mysecondaryaddress]{Songcan
Chen\corref{mycorrespondingauthor}}
\cortext[mycorrespondingauthor]{Corresponding author}
\ead{s.chen@nuaa.edu.cn}

\address[mymainaddress]{College of Computer Science \& Technology, Nanjing University of Aeronautics \& Astronautics,
Nanjing, Jiangsu, 211106, China}
\address[mysecondaryaddress]{MIIT Key Laboratory of Pattern Analysis and Machine Intelligence, Nanjing University of Aeronautics \& Astronautics, Nanjing, Jiangsu, 211106, China}

\begin{abstract}
Unsupervised Domain Adaptation (UDA) aims to classify unlabeled target domain by transferring knowledge from labeled source domain with domain shift. Most of the existing UDA methods try to mitigate the adverse impact induced by the shift via reducing domain discrepancy. However, such approaches easily suffer a notorious mode collapse issue due to the lack of labels in target domain. Naturally, one of the effective ways to mitigate this issue is to reliably estimate the pseudo labels for target domain, which itself is hard. To overcome this, we propose a novel UDA method named Progressive Adaptation of Subspaces approach (PAS) in which we utilize such an intuition that appears much reasonable to gradually obtain reliable pseudo labels. Specifically, we \emph{progressively and steadily refine} the shared subspaces as bridge of knowledge transfer by adaptively anchoring/selecting and leveraging those target samples with reliable pseudo labels. Subsequently, the refined subspaces can in turn provide more reliable pseudo-labels of the target domain, making the mode collapse highly mitigated. Our thorough evaluation demonstrates that PAS is not only effective for common UDA, but also outperforms the state-of-the arts for more challenging Partial Domain Adaptation (PDA) situation, where the source label set subsumes the target one.
\end{abstract}


\end{frontmatter}

\section{Introduction}
\begin{figure}[t]
    \centering
    \subfigure[]{
        \includegraphics[width=1\textwidth]{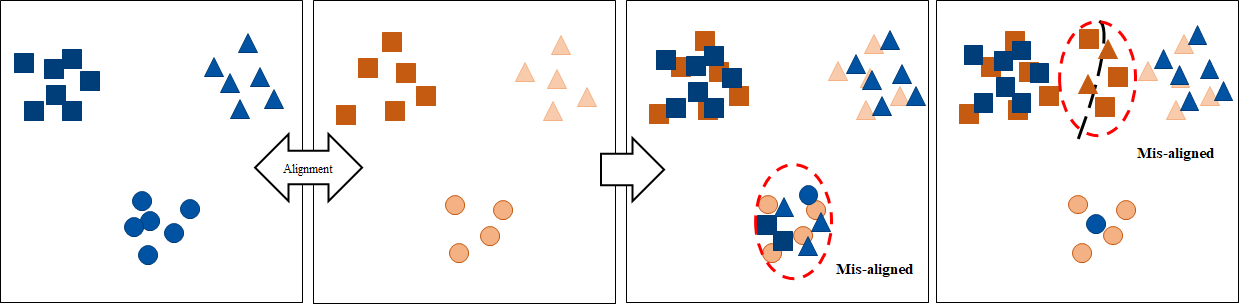}
        \label{figure1a}
    }\\
    \subfigure[]{
        \includegraphics[width=1\textwidth]{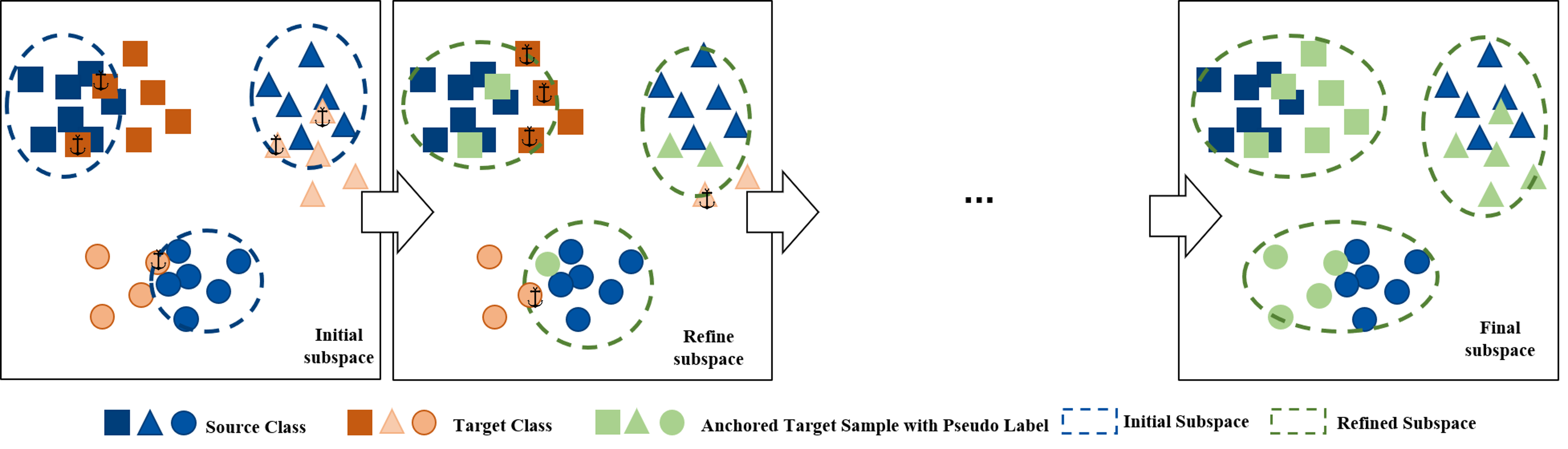}
        \label{figure1b}
    }
    \caption{(a).Existing approaches easily suffer the notorious mode collapse issue induced by aligning the unknown target domain. (b) The framework of proposed PAS approach: we progressively anchor target samples with reliable pseudo label for refining the corresponding subspaces.}
    \label{fig:fig_micPerMon}
\end{figure}

Machine learning approaches have achieved impressive results on several applications with sufficient labeled training datasets. Unfortunately, due to the phenomenon known as domain shift \cite{kouw2019review}, a well pre-trained model in some realistic applications often generalizes poorly on related target domain with different distribution  \cite{pan2010domain,fernando2013unsupervised,ganin2015unsupervised,long2013adaptation,tsuboi2009direct}.
Besides, it is often too expensive and time-consuming to re-collect massive amounts of labeled data. As a promising strategy, UDA effectively enhances the generalization from labeled source domain to unlabeled target domain by intentionally alleviating domain shift \cite{ben2010theory}. Currently, there have had various UDA methods proposed to mitigate the adverse impact of shift \cite{kouw2019review}, which can be summarized into the following three main paradigms: (1) sample-based paradigm, which corrects the shift by importance-weighting of samples \cite{tsuboi2009direct,sugiyama2008direct,huang2007correcting}; (2) model-based paradigm, which adapts the model parameter of source domain to target domain by imposing additional constraint \cite{long2013adaptation,baktashmotlagh2013unsupervised,belkin2006manifold}; and (3) subspace-based paradigm, which learns the common or domain-invariant subspaces by distribution or subspace alignment \cite{pan2010domain,fernando2013unsupervised,ganin2015unsupervised}.\\
Unfortunately, although all three above-mentioned paradigms can reduce the domain discrepancy effectively, they still suffer the notorious mode collapse issue induced by aligning the unknown target domain with multimodal structures/distribution \cite{long2018conditional}, as shown in Figure \ref{figure1a}, which limits their performance, even the label sets of the domains involved are all identical \cite{kurmi2019looking,tang2020discriminative}.  In particular, when the label set of all target domain just corresponds to a subset of the source one, forming a so-called partial domain adaptation (PDA) \cite{cao2018partial}, such issues will further drastically aggravate the negative transfer due to the existence of irrelevant or extra classes. Most of the existing approaches utilize the pseudo labels of target samples to alleviate this issue respectively by fine-grained alignment \cite{pei2018multi}, instance reweighting \cite{cao2018partialCVPR,long2014transfer} or discriminative information incorporation \cite{long2018conditional,tang2020discriminative}. Very naturally, the reliability of so-estimated pseudo labels of target domain becomes a key to mitigate such a collapse. However, current strategies are still relatively vulnerable due to that the target samples are easily misclassified invoked by the shift and mode collapse. \\
To address this problem, a practicable alternative is to progressively anchor the target samples which have relatively reliable pseudo labels for adaptation. The intuition is that those pseudo labels with gradually increasing confidence in learning appear to be more reliable. Therefore, as a try, we propose a novel UDA method named Progressive Adaptation of Subspaces approach (PAS) where we follow the subspace-based paradigm mainly for its simplicity. \\
Specifically, we assume that both source and target samples share the $K$ common subspaces corresponding to $K$ classes for the multi-modal structure of data \cite{cao2019learning}. Then, the knowledge can be transferred by learning the shared subspaces between the domains involved. Since the target labels are unknown, we first need to learn the initial subspaces with the source domain. After that, we progressively anchor and leverage the target samples with reliable pseudo labels to refine the shared subspaces for adaptation, since the support for the initial subspaces are defined only over the source domain instances. Subsequently, the refined subspaces in turn improve the reliability of the pseudo labels for target samples, allowing us to anchor more samples until traversing all target samples. In this way, we considerably mitigate the risk of mode collapse in adaptation with the reliable pseudo labels. \\
The main idea of PAS is shown in Figure \ref{figure1b}. For facilitating efforts to replicate our results, our implementation is available on GitHub \footnote{\href{https://github.com/Cavin-Lee/PAS}{https://github.com/Cavin-Lee/PAS}.}.
In summary, this work makes the following contributions:
\begin{enumerate}
    \item We explore a novel UDA method named Progressive Adaptation of Subspaces (PAS) for effectively alleviating the mode collapse in domain adaptation.
    \item We provide an effective algorithm to implement PAS, which progressively anchors and leverages the target samples with reliable pseudo labels to refine the shared subspaces.
    \item Experimental results on several public datasets demonstrate the effectiveness of PAS, especially on a more realistic and challenging scenario (i.e., partial domain adaptation).
\end{enumerate}
The rest of this paper is organized as follows. In Section \ref{s1}, we briefly overview unsupervised domain adaptation and partial domain adaptation. In Section \ref{s3}, we elaborate on the problem formulation, PAS model and optimizing algorithm in detail. The experimental results and the analysis are reported in Section \ref{s4}. In the end, we conclude the entire paper with future research directions in Section \ref{s5}.
\section{Related Work\label{s1}}
\subsection{Unsupervised domain adaptation}
The common setting for UDA is also known as closed-set UDA, where source and target domains share the identical label set \cite{kouw2019review}. Several methods are proposed for close-set UDA. In the present study, we focus on subspace-based UDA paradigm, most of which can be divided in two categories as follows.
\paragraph{Distribution alignment} This category tries to directly align the source and target domains by learning the joint latent/explicit subspaces to minimize a certain domain dissimilarity measure between the source and target domain, as shown in Eq.(\ref{eq1}).
\begin{equation}
    \label{eq1}
    \begin{split}
       \mathop{\min}_{\mathbf{U}}F(\mathbf{U})+&\lambda D(\mathcal{D}_s\|\mathcal{D}_t)\\
       s.t.,\ \mathbf{U} &\in \Omega
    \end{split}
\end{equation} 
where $D(\mathcal{D}_s\|\mathcal{D}_t)$ is a domain dissimilarity measure between source domain $\mathcal{D}_s$ and target domain $\mathcal{D}_t$ on the learned subspaces $\mathbf{U}$,  such as  Maximum Mean Discrepancy \cite{pan2010domain}, Joint Maximum Mean Discrepancy  \cite{long2013adaptation,long2017deep}, Maximum Classifier Discrepancy \cite{saito2018maximum},  Central Moment Discrepancy \cite{zellinger2017central}, Bregman divergence \cite{si2009bregman},Wasserstein distance \cite{courty2014domain}, $\mathcal{A}$-distance \cite{ganin2015unsupervised}, $\Delta$-distance \cite{long2018conditional} or the distance between the second-order statistics (covariance) of the source and target features \cite{sun2016return}. $F(\mathbf{U})$ is the objective function of the subspace learning, which can be formulated in deep or shallow format. $\Omega$ is the constraint for subspace. 
\paragraph{Subspace alignment} This category assumes that the domains contain domain-specific noise but common subspaces. Consequently, the goal of these methods is to find the optimal transform of the subspaces for matching the source data to the target data, which can be summarized as Eq.(\ref{eq2}):
\begin{equation}
    \label{eq2}
    \mathop{\min}_{T(\cdot)}D(T(\mathbf{U}_s)\|\mathbf{U}_t)
\end{equation}
where $T(\cdot)$ is a transformation operator to transform the bases of single or multiple subspaces from source $\mathbf{U}_s$ to target domains $\mathbf{U}_t$ (or vice versa) \cite{thopalli2019multiple}, such as linear transformation  \cite{fernando2013unsupervised} or  Geodesic Flow Kernel  \cite{gong2012geodesic}, etc. The commonly-used domain dissimilarity measure for subspace alignment contains Hellinger distance \cite{gong2012geodesic} or Target density around source (TDAS) \cite{fernando2013unsupervised}, etc.\\
The existing subspace-based UDA methods mainly focus on directly learning the shared subspaces to transfer the classifier/knowledge obtained in the source domain. However, these methods still suffer the  potential risk of negative transfer, due to the mode collapse, especially for the partial domain adaptation, whose details will be further discussed in next section. Besides,  such methods commonly learn a classifier only on source domain without exploiting any information from target domain, which may limit its performance.\\
Different from these methods which directly align the entire subspaces/distributions between source and target domains, in this work,  to alleviate the mode collapse in adaptation, we progressively refine the shared common subspaces by adaptively anchoring target samples with reliable pseudo labels. In addition, the  subspaces learned by PAS also play a role of classifier, and are learned from both source and target domains. In other word, PAS can effectively utilize the discriminative information from both source and target domains.
\subsection{Partial Domain Adaptation}
Since the label set of target domain tends to be unknown, it is hard to find a proper source domain with identical label set to the target domain. Thus, a more realistic scenario is to transfer knowledge from existing rich domain with sufficiently large label set (e.g., ImageNet \cite{Russakovsky2015ImageNet}) to unlabeled target domain. As a result, the target domain may not include some classes present in the source domain, which is contrary to the assumption of closed-set UDA. Such scenario naturally derives a new setting for DA, namely partial domain adaptation (PDA), which assumes that the target label space is a subspace of the source label space \cite{cao2018partial}. Due to the existence of the irrelevant categories in source domain, the risk of mode collapse in PDA is highly increased in alignment. To solve PDA, the existing approaches mainly focus on a reweighting strategy on example-level \cite{cao2018partial,cao2019learning,zhang2018importance} or class-level \cite{cao2018partialCVPR,matsuura2018twins} to mitigate the potential negative transfer caused by irrelevant examples or classes.\\
Although these  approaches achieve performance gains over standard closed-set UDA methods in PDA tasks. Such methods do not fundamentally solve the problem of more essential mode collapse. Moreover, such methods commonly cannot work well on closed-set UDA due to their special design for the PDA setting. In contrast, the proposed PAS approach focuses on mitigating the risk of mode collapse in adaptation by gradually providing reliability-increasing pseudo-labels of the target domain while progressively refining the shared subspaces. In this way, PAS is naturally suitable for both closed-set UDA and PDA without additional specially designed reweighting strategy.
\section{Progressive Adaptation of Subspaces\label{s3}}
In this section, we go through the details of the proposed Progressive Adaptation of Subspaces (PAS). In what follows, we detail the formulation of problem, each component and the learning algorithm for PAS.
\subsection{Problem Definition}
Given a labeled source domain $\mathcal{D}_s=\{ \mathbf{x}_i^s,y_i^s\}_{i=1}^n$ with $n$ samples and an unlabeled target domain $\mathcal{D}_t=\{ \mathbf{x}_i^t\}_{i=1}^m$ with $m$ samples, where their features $\mathbf{x}_i^s,\mathbf{x}_i^t \in \mathbb{R}^d$. The source and target domains follow the distributions $p$ and $q$, respectively, and $p \ne q$. Both domains are assumed from $K$ subspaces corresponding to the label space $\{1,2,…,K\}$, where $K$ is the number of classes. Specifically, for the setting of the closed-set UDA, source and target domains share the identical label set. As for PDA, the label set of target domain is a subset of source one. Thus, we further have $p_C \ne q$ in PDA, where $p_C$ denotes the distribution of samples belonging to the common categories in source domain, but the common categories are unknown. The ultimate goal of PAS is to learn the shared $K$ multiple subspaces for knowledge transferring based on the labeled source and the unlabeled target domains. 
\subsection{Progressive Adaptation}
In this paper, we propose to alleviate the mode collapse  in learning by progressive adaptation so that the shared subspaces $\mathbf{U}=\{\mathbf{U}_1,...,\mathbf{U}_K\}$ can progressively be refined through gradually anchoring and leveraging target samples with reliable pseudo labels. To achieve this, the progressive adaptation involves two key factors, i.e., subspace learning and sample anchoring. \\
Formally, we denote $\mathbf{U}_k \in \mathbb{R}^{d\times D}$ as the base for the subspace $k$ with $D$ dimension and $\mathbf{v}=[\mathbf{v}_{j}]_{m}$ as the indicator index of the anchored target samples. Then, the objective function of implementing the progressive adaptation can be formulated as:
\begin{equation}
    \label{eq3}
    \begin{split}
       \mathop{\min}_{\mathbf{U},\mathbf{W},\mathbf{v}}\sum_{i=1}^n\sum_{k=1}^K\mathbf{Y}_{ik}D(\mathbf{x}_i^s,\mathbf{U}_k) +\sum_{j=1}^m\mathbf{v}_j&\sum_{k=1}^K\mathbf{W}_{jk}D(\mathbf{x}_j^t,\mathbf{U}_k)+\lambda R(\mathbf{v})\\
       s.t.,\mathbf{U}_k\in \Delta,\mathbf{v}_j\in\{0&,1\}, \mathbf{W}_{jk}\in\{0,1\},\\
       \sum_{k=1}^K\mathbf{W}_{jk}=1,\forall j=1&\dots m, k=1\dots K
    \end{split}
\end{equation} 
where $D(\mathbf{x},\mathbf{U}_k)$ is the distance from $\mathbf{x}$ to $\mathbf{U}_k$ for learning the shared subspaces,  $R(\mathbf{v})$ refers to the regularization term for sample anchoring, $\lambda$ is the trade-off hyper-parameter for controlling the anchored sample size.  $\Delta$ is the constraint of subspace learning such as orthogonal constraint. $\mathbf{W}=[\mathbf{W}_{jk}]_{m \times  K}$ denotes the predictive membership matrix, in which $\mathbf{W}_{jk}$ represents the predictive membership of $\mathbf{x}_j^t$ to the $k$th subspace. Similarly, $\mathbf{Y}=[\mathbf{Y}_{ik}]_{n \times  K}$ denotes the membership matrix of source samples, in which $\mathbf{Y}_{ik}$ represents the predictive membership of $\mathbf{x}_i^s$ to the $k$th subspace.\\
In practice, the target samples with $\mathbf{v}_j=1$ will be anchored to refine the corresponding subspace $\mathbf{U}_k$  with $\mathbf{W}_{jk}=1$. By updating the hyper-parameter $\lambda$, more target samples can adaptively be anchored to progressively refine the shared subspaces. Meanwhile, the refined subspaces in turn improve the reliability of the pseudo labels or $\mathbf{W}$, since the estimation of $\mathbf{W}$ highly relies on the distance between target sample and its corresponding subspace. \\
In next sections, we first introduce subspace learning (i.e., $D(\mathbf{x},\mathbf{U}_k)$). Then, we present the  sample anchoring (i.e., $R(\mathbf{v})$). Eventually, we articulate the unified objective function of PAS.
\subsection{Subspace Learning}
The purpose of subspace learning is to learn the shared subspace as a bridge for knowledge transfer. In principle, any subspace learning objective can all be adopted to realize the PAS. In this paper, referring to the $K$-subspace learning \cite{agarwal2004k}, we select the squared distance as $D(\mathbf{x},\mathbf{U}_k)$ due to its simplicity and intuitiveness \cite{pearson1901principal}. Then, the objective function for subspace learning is formulated as follows. 
\begin{equation}
    \label{eq4}
    \begin{split}
       \mathop{\min}_{\mathbf{U},\mathbf{Z},}\sum_{i=1}^n\sum_{k=1}^K\mathbf{Y}_{ik}\|\mathbf{x}_i^s-\mathbf{U}_k\mathbf{Z}_{si}^k\|_F^2 +&\sum_{j=1}^m \mathbf{v}_j\sum_{k=1}^K\mathbf{W}_{jk}\|\mathbf{x}_j^t-\mathbf{U}_k\mathbf{Z}_{tj}^k\|_F^2\\
       s.t.\mathbf{U}_k^T\mathbf{U}_k=\mathbf{I},&\forall k=1\dots K
    \end{split}
\end{equation} 
where $\mathbf{I}$ is the identity matrix.  $\mathbf{Z}_{si}^k \in \mathbb{R}^{D\times n}$ and $\mathbf{Z}_{tj}^k  \in \mathbb{R}^{D\times m}$ are the representations of source sample $\mathbf{x}_i^s$ and  target sample $\mathbf{x}_j^t$ on the subspace $\mathbf{U}_k$, respectively. With  $\mathbf{Y}$, $\mathbf{v}$ and $\mathbf{W}$ given, Eq.(\ref{eq4}) decouples as the sum of $K$ objective functions, per subspace. Since each objective function is identical to that minimized by standard PCA problem \cite{vidal2005generalized}, the optimal $\{\mathbf{U}_k\}_{k=1}^K$ can easily be solved by applying PCA to each subspace. 
\subsection{Sample Anchoring}
One crucial problem for PAS is how to anchor or select the target samples which have relatively more reliable pseudo labels. Motivated by the DA theory of Ben David \cite{ben2010theory}, an intuitive alternative is to anchor the target samples close to the source domain.  Since the shared subspaces are supported by source domain, we progressively anchor the target samples close to the shared subspaces by terming negative $l_1$-norm regularizer as $R(\mathbf{v})$,  expressed as:
\begin{equation}
    \label{eq5}
    \begin{split}
       \mathop{\min}_{\mathbf{v}}\sum_{j=1}^m\mathbf{v}_j\sum_{k=1}^K\mathbf{W}_{jk} & D(\mathbf{x}_j^t,\mathbf{U}_k)-\lambda \sum_{j=1}^m\mathbf{v}_j\\
       s.t.\mathbf{v}_j \in \{0,1\},&\forall j=1\dots m
    \end{split}
\end{equation} 
In Eq.(\ref{eq5}), when the value of $\lambda$ approaches 0, only few target samples will be anchored. In contrast, when $\lambda$ has a large value, almost all target samples will be anchored. In other words, with the increase of hyper-parameter $\lambda$, more target samples can gradually be anchored to progressively refine the shared subspaces. In addition, the proposed sample anchoring approach in Eq.(\ref{eq5}) can highly reduce the risk of mode collapse in subspaces learning, since target samples with unreliable pseudo-labels tend to not be anchored due to its large distance to the subspaces, which is also conformed in our experiment later.
\subsection{Unified Objective Function}
By simply integrating Eq.(\ref{eq4}) as well as Eq.(\ref{eq5}) into Eq.(\ref{eq3}), we obtain the following PAS optimization problem: 
\begin{equation}
    \label{eq6}
    \begin{split}
       \mathop{\min}_{\mathbf{U},\mathbf{W},\mathbf{Z},v}\sum_{i=1}^n\sum_{k=1}^K\mathbf{Y}_{ik}\|&\mathbf{x}_i^s-\mathbf{U}_k\mathbf{Z}_i^k\|_F^2 +\\
       \sum_{j=1}^m \mathbf{v}_j\sum_{k=1}^K\mathbf{W}_{jk}\|\mathbf{x}_j^t-&\mathbf{U}_k\mathbf{Z}_j^k\|_F^2 -\lambda \sum_{j=1}^m\mathbf{v}_j\\
       s.t.\mathbf{U}_k^T\mathbf{U}_k=\mathbf{I},\mathbf{v}_j\in&\{0,1\},\mathbf{W}_{jk}\in \{0,1\},\\
       \sum_{k=1}^K\mathbf{W}_{jk}=1,\forall j=1&\dots m,k=1\dots K
    \end{split}
\end{equation} 
According to Eq.(\ref{eq6}), the target samples with low  squared distance are selected to refine the shared subspaces. Note that, since the target labels are unknown, we initialize the subspaces by setting  $\lambda=0$, and Eq.(\ref{eq6}) returns a trivial solution that $\mathbf{v}_j = 0$ for all $j=\{1,\dots ,m\}$, meaning that no target samples are anchored in the first iteration. Then,  we gradually increase the hyper-parameter $\lambda$ to incorporate target samples from 0\%  to 100 \%  of target domain with 1\% step length to progressively refine the shared subspaces. 
\subsection{Optimization}
The objective function in Eq.(\ref{eq6}) is not jointly convex in all variables. To progressively anchor the target samples and refine the shared subspaces, we iteratively update $\mathbf{v}$, $\mathbf{W}$, $\mathbf{U}$ and $\mathbf{Z}$ with the gradually increasing  hyper-parameter $\lambda$, whose details are as follows.\\
\noindent
\textbf{Updating} $\mathbf{U}_k$, $\mathbf{Z}_s^k$ and $\mathbf{Z}_t^k$, with $\mathbf{W}$ and $\mathbf{v}$ fixed, Eq.(\ref{eq6}) is degenerated to the following optimization problem.
\begin{equation}
    \label{eq7}
    \begin{split}
       \mathop{\min}_{\mathbf{U}_k,\mathbf{Z}_s^k,\mathbf{Z}_t^k}\sum_{i=1}^n\sum_{k=1}^K\mathbf{Y}_{ik}\|\mathbf{x}_i^s-\mathbf{U}_k\mathbf{Z}_{si}^k\|_F^2 +&\sum_{j=1}^m \mathbf{v}_j\sum_{k=1}^K\mathbf{W}_{jk}\|\mathbf{x}_j^t-\mathbf{U}_k\mathbf{Z}_{tj}^k\|_F^2\\
       s.t.\mathbf{U}_k^T\mathbf{U}_k=\mathbf{I},&\forall k=1\dots K
    \end{split}
\end{equation} 
As mentioned before, $\mathbf{U}_k$ can be easily solved by PCA.
\begin{equation}
    \label{eq8}
    \begin{split}
       \mathbf{U}_k &= \mathbf{\mathcal{U}}_{1:D}\\
       s.t.\mathbf{\Lambda}\mathcal{U}&=\mathbf{X}_k^T\mathbf{H}\mathbf{X}_k\mathbf{\mathcal{U}}
    \end{split}
\end{equation} 
where $\mathbf{\mathcal{U}}_{1:D}$ is the rank-$D$ columns of matrix $\mathbf{\mathcal{U}}$, $\mathbf{H}$ is the centering matrix and $\mathbf{X}_k$ is data matrix of the source and target data with $\mathbf{W}_{jk}=1,\mathbf{v}_j=1$ or  $\mathbf{Y}_{ik}=1$.\\
Then, based on the least square method\cite{aldroubi2009nonlinear}, we can easily get the closed-form solutions for $\mathbf{Z}_s^k$ and $\mathbf{Z}_t^k$:
\begin{equation}
    \label{eq9}
    \begin{split}
       \mathbf{Z}_s^k=\mathbf{U}_k^T\mathbf{X}_s\\
       \mathbf{Z}_t^k=\mathbf{U}_k^T\mathbf{X}_t
    \end{split}
\end{equation} 
\textbf{Updating} $\mathbf{W}$, with $\mathbf{U}$, $\mathbf{Z}$ and $\mathbf{v}$ fixed, Eq.(\ref{eq6}) turns to the following minimization problem:
\begin{equation}
    \label{eq10}
    \begin{split}
       \mathop{\min}_{\mathbf{W}_{jk}}\sum_{j=1}^m \mathbf{v}_j\sum_{k=1}^K\mathbf{W}_{jk}&\|\mathbf{x}_j^t-\mathbf{U}_k\mathbf{Z}_j^k\|_F^2\\
       s.t.\mathbf{W}_{jk}\in \{0,1\},\sum_{k=1}^K\mathbf{W}_{jk}=1,&\forall j=1\dots m,k=1\dots K
    \end{split}
\end{equation} 
We can easily get the optimal solution of $\mathbf{W}$ as follow:
\begin{equation}
    \label{eq11}
    \mathbf{W}_{jk}=
    \begin{cases}
       1& if\ k =\mathop{{\arg\min}}_k\|\mathbf{x}_j^t-\mathbf{U}\mathbf{Z}^k\|_F^2\\
       0& \quad\quad\quad else
    \end{cases}
\end{equation} 
\textbf{Updating} $\mathbf{v}$, with$\mathbf{W}$, $\mathbf{U}$ and $\mathbf{Z}$ fixed, we need to minimize the following objective function:
\begin{equation}
    \label{eq12}
    \begin{split}
       \mathop{\min}_{\mathbf{v}_j}\sum_{j=1}^m \mathbf{c}_j\sum_{k=1}^K\mathbf{W}_{jk}\|\mathbf{x}_j^t&-\mathbf{U}_k\mathbf{Z}_j^k\|_F^2-\lambda \sum_{j=1}^m\mathbf{v}_j\\
       s.t.\mathbf{v}_j\in\{0,1\},&\forall j=1\dots m
    \end{split}
\end{equation} 
Unfortunately, Optimizing $\mathbf{v}_j$ in the Eq.(\ref{eq12}) is NP hard \cite{kumar2010self}. To solve $\mathbf{v}_j$, we relax $\mathbf{v}_j$ to the interval $[0,1]$.
\begin{equation}
    \label{eq13}
    \begin{split}
       \mathop{\min}_{\mathbf{v}_j}\sum_{j=1}^m \mathbf{v}_j\sum_{k=1}^K\mathbf{W}_{jk}\|x_j^t&-\mathbf{U}_k\mathbf{Z}_j^k\|_F^2-\lambda \sum_{j=1}^m\mathbf{v}_j\\
       s.t.\mathbf{v}_j\in[0,1],&\forall j=1\dots m
    \end{split}
\end{equation} 
which can be further simplified to the following problem
\begin{equation}
    \label{eq14}
    \begin{split}
       \mathop{\min}_{\mathbf{v}_j}\sum_{j=1}^m &\mathbf{v}_j(\mathbf{c}_j-\lambda)\\
       s.t.\mathbf{v}_j\in[0,1],&\forall j=1\dots m
    \end{split}
\end{equation} 
where $\mathbf{v}_j =\sum_{k=1}^K\mathbf{W}_{jk}\|\mathbf{x}_j^t-\mathbf{U}_k\mathbf{Z}_j^k\|_F^2$ (i.e., distance to the its own subspace) is constant here. Note that Eq.(\ref{eq14}) reduces into a simple linear programming problem, and thus we can easily write the optimal solution for $\mathbf{v}_j$ as follows
\begin{equation}
    \label{eq15}
    \mathbf{v}_j=
    \begin{cases}
       1& if\ \sum_{k=1}^K\mathbf{W}_{jk}\|\mathbf{x}^t-\mathbf{U}_k\mathbf{Z}_j^k\|_F^2<\lambda\\
       0& \quad\quad\quad else
    \end{cases}
\end{equation}  
Consequently, the relaxation is tight. The entire optimization procedure for PAS is summarized in Algorithm \ref{alg1}
\begin{algorithm}[ht]
    \caption{Learning algorithm for PAS}
    \label{alg1} 
    \begin{algorithmic}[1]
    \REQUIRE ~~\\
    $n$ source labeled datasets $\mathcal{D}_s=\{\mathbf{x}^s_i,y_i\}_{i=1}^n$\\
    $m$ target unlabeled datasets $\mathcal{D}_t=\{\mathbf{x}^s_i\}_{i=1}^m$\\
    Dimension of subspaces $D$;\\ 

    \ENSURE ~~\\ 
    Predictive membership matrix $\mathbf{W}$
    \WHILE {not converge}
    \STATE Update  $\mathbf{U}_k$  by Eq.(\ref{eq8});
    \STATE Update  $\mathbf{Z}_s^k$ and $\mathbf{Z}_t^k$ by Eq.(\ref{eq9});
    \STATE Update  $\mathbf{W}$ by Eq.(\ref{eq11});
    \STATE Update  $\mathbf{v}$ by Eq.(\ref{eq15});
    \STATE \textbf{IF}  $\lambda$  is small \textbf{THEN} increase $\lambda$ by the step size;
    \ENDWHILE 
    \STATE \textbf{Return} $\mathbf{W}$ and $\mathbf{U}$.
    \end{algorithmic}
\end{algorithm}
 \subsection{Analysis}
\subsubsection{Complexity Analysis}
The time complexity of PAS mainly consists of the complexities of solving all the variables. We denote $T$ as the number of iterations, $K$ as the number of classes, $D$ as the dimension of the shared subspace and $d$ as the dimension of input features. In each iteration, for updating $\{\mathbf{U}_k\}_{k=1}^K$, the  major time-consuming terms is PCA procedure, which needs $\mathcal{O}(KDd^2+Kdn^2)$.  For updating $\{\mathbf{Z}_t\}_{k=1}^K$, it contains $K$ matrix multiplication and inverse operations, thus costs $\mathcal{O}(Kd^3)$.  For updating $\mathbf{W}$, it contains $K$ traverse operation, and the complexity is $\mathcal{O}(Kn)$. For updating $\mathbf{v}$, it contains a traverse operation and the complexity is $\mathcal{O}(n)$. Generally, we have $D\ll d$ and $n\ll d^3 $, so the total complexity of PAS is  $\mathcal{O}(TKd^3+TKdn^2)$.
\subsubsection{Convergence Analysis}
Denote by  $\mathcal{J}(\mathbf{U},\mathbf{Z},\mathbf{W}, \mathbf{v})$ the objective function value of Eq.(\ref{eq6}). The Algorithm \ref{alg1} monotonically decreases the objective value of the problem in Eq.(\ref{eq6}) in each iteration. To prove it, we need the following lemmas:

\begin{lemma}
\label{lem1}
Let $\mathbf{L}\in \mathbb{R}^{m\times K}$ be a given matrix. If the optimal solution to
\begin{equation}
    \label{eqlem}
    \begin{split}
       \mathop{\min}_{\mathbf{W}}\sum_{j=1}^m \sum_{k=1}^K&\mathbf{W}_{jk}\mathbf{L}_{jk}\\
       s.t.\mathbf{W}_{jk}\in \{0,1\},\sum_{k=1}^K\mathbf{W}_{jk}=1,&\forall j=1\dots m,k=1\dots K
    \end{split}
\end{equation}    
is $\mathbf{W}^*$, then the  $\mathbf{W}_{jk}^*$ is:
\end{lemma}
\begin{equation}
    \label{eqbb}
    \mathbf{W}^*_{jk}=
    \begin{cases}
       1& if\ k =\mathop{{\arg\min}}_k \mathbf{L}_{jk}\\
       0& \quad\quad\quad else
    \end{cases}
\end{equation} 
\noindent
\textbf{Proof} Lemma \ref{lem1} can easily be proved by contradiction: we first assume that $\hat{\mathbf{W}} \neq \mathbf{W}^*$ is the optimal solution to Eq.(\ref{eqlem}). Let $\mathbf{L}_{j*}$ be the minimum value as the $j$-th column of $\mathbf{L}$ and $\hat{\mathbf{L}}_{j*}$ be the value of $\mathbf{L}_{jq}$, where $\hat{\mathbf{W}}_{jq}=1$. Then, we have:
\begin{equation}
    \begin{split}
   \sum_{j=1}^m \sum_{k=1}^K\hat{\mathbf{W}}_{jk}\mathbf{L}_{jk}=\hat{\mathbf{L}}_{1*}+\hat{\mathbf{L}}_{2*}+\dots+\hat{\mathbf{L}}_{m*}\\
   \leq\sum_{j=1}^m \sum_{k=1}^K\mathbf{W}^*_{jk}\mathbf{L}_{jk}=\mathbf{L}_{1*}+L_{2*}+\dots+\mathbf{L}_{m*},
    \end{split}
\end{equation}
which can be reformulated to:
\begin{equation}
  \label{eqp1}
   (\hat{\mathbf{L}}_{1*}-\mathbf{L}_{1*})+(\hat{\mathbf{L}}_{2*}-\mathbf{L}{2*})+\dots+(\hat{\mathbf{L}}_{m*}-\mathbf{L}_{m*})\leq 0.
\end{equation} 
According to the definition of $\mathbf{L}_{j*}$, $\forall j\in 1 \dots m$ we easily have $\hat{\mathbf{L}}_{j*}-\mathbf{L}_{j*}\geq0$, and the equality arrives when $\mathbf{W}_{jk}^*=\hat{\mathbf{W}_{jk}}$. Since $\hat{\mathbf{W}} \neq \mathbf{W}^*$, Eq.(\ref{eqp1}) does not hold, which contradicts the assumption that $\hat{\mathbf{W}}$ is the optimal solution. Thus, $\mathbf{W}^*$ is the optimal solution of Eq.(\ref{eqlem}). 
\begin{lemma}
\label{lem2}
 Let $\mathbf{c}\in \mathbb{R}^{m}$ be a given vector and $\lambda$ be a constant. If the optimal solution to
\begin{equation}
    \label{eqlem2}
    \begin{split}
       \mathop{\min}_{\mathbf{v}}\sum_{j=1}^m &\mathbf{v}_j(\mathbf{c}_j-\lambda)\\
       s.t.\mathbf{v}_j\in\{0,1\},&\forall j=1\dots m
    \end{split}
\end{equation}     
is $\mathbf{v}^*$, then the $\mathbf{v}_{j}^*$ is:
\end{lemma}
\begin{equation}
    \label{eqe22}
     \mathbf{v}_j=
    \begin{cases}
       1& if\quad \mathbf{c}_j<\lambda\\
       0& \quad\quad else
    \end{cases}
\end{equation}  
\noindent
\textbf{Proof} Lemma \ref{lem2} can easily be proved by contradiction: we assume that $\hat{\mathbf{v}} \neq \mathbf{v}^*$ is the optimal solution to Eq.(\ref{eqlem2}), we have 
\begin{equation}
   \sum_{j=1}^m \hat{\mathbf{v}}_j(\mathbf{c}_j-\lambda) \leq \sum_{j=1}^m \mathbf{v}^*_j(\mathbf{c}_j-\lambda),
\end{equation}
which can be reformulated to:
\begin{equation}
\label{eqlem3}
   \begin{split}
   \sum_{j=1}^m (\hat{\mathbf{v}}_j-\mathbf{v}^*_j)\times(\mathbf{c}_j-\lambda) \quad= & \sum_{\hat{\mathbf{v}}_j=1,\mathbf{v}^*_j=0}(\mathbf{c}_j-\lambda)\quad-\sum_{\hat{\mathbf{v}}_j=0,\mathbf{v}^*_j=1}(\mathbf{c}_j-\lambda)\\
   \leq& \quad 0
   \end{split}
\end{equation}
However, based on Eq.(\ref{eqe22}), we easily have $\sum_{\hat{\mathbf{v}}_j=1,\mathbf{v}^*_j=0}(\mathbf{c}_j-\lambda)-\sum_{\hat{\mathbf{v}}_j=0,\mathbf{v}^*_j=1}(\mathbf{c}_j-\lambda)\geq 0$, which is contradicts with the Eq.(\ref{eqlem3}). Thus, $\mathbf{v}^*$ is the optimal solution of Eq.(\ref{eqlem2}). Then, the convergence of the Algorithm \ref{alg1} is summarized in the following:
\begin{theorem}
\label{the1}
 The objective function value shown in Eq.(\ref{eq6}) monotonically decreases until convergence by applying the proposed algorithm.
\end{theorem}
\noindent
\textbf{Proof} Suppose after  $t$-th iteration, we have $\mathcal{J}(\mathbf{U}^{(t)},\mathbf{Z}^{(t)},\mathbf{W}^{(t)}, \mathbf{v}^{(t)})$ as the objective function value. In the next iteration, we fix $\mathbf{W}^(t)$ and  $\mathbf{v}^(t)$. Then, the objective function decouples as the sum of $K$ PCA objective functions. Thus, it can be easily verified that $\mathbf{U}^{(t+1)}$ and $\mathbf{Z}^{(t+1)}$ are the global optimum for $\mathcal{J}(\mathbf{U},\mathbf{Z},\mathbf{W}^{(t)}, \mathbf{v}^{(t)})$ \cite{agarwal2004k,pearson1901principal}, and we have:
\begin{equation} \label{eqt1}
    \mathcal{J}(\mathbf{U}^{(t+1)},\mathbf{Z}^{(t+1)},\mathbf{W}^{(t)}, \mathbf{v}^{(t)})\leq \mathcal{J}(\mathbf{U}^{(t)},\mathbf{Z}^{(t)},\mathbf{W}^{(t)}, \mathbf{v}^{(t)})
\end{equation}
For the variable $\mathbf{W}$, resorting to the Lemma \ref{lem1}, we have:
\begin{equation}
    \begin{split}
    \sum_{j=1}^m\sum_{k=1}^K\mathbf{W}_{jk}^{(t+1)}\|\mathbf{x}_j^t-\mathbf{U}_k^{(t+1)}\mathbf{Z}_j^{k(t+1)}\|_F^2\\
    \leq\sum_{j=1}^m\sum_{k=1}^K\mathbf{W}_{jk}^{(t)}\|\mathbf{x}_j^t-\mathbf{U}_k^{(t+1)}\mathbf{Z}_j^{k(t+1)}\|_F^2
    \end{split}
\end{equation}
Thus, $\mathbf{W}^{(t+1)}$ is the global optimum for $\mathcal{J}(\mathbf{U}^{(t+1)},\mathbf{Z}^{(t+1)},\mathbf{W}, \mathbf{v}^{(t)})$ and the following inequality holds:
\begin{equation}\label{eqt2}
    \mathcal{J}(\mathbf{U}^{(t+1)},\mathbf{Z}^{(t+1)},\mathbf{W}^{(t+1)}, \mathbf{v}^{(t)})\leq \mathcal{J}(\mathbf{U}^{(t+1)},\mathbf{Z}^{(t+1)},\mathbf{W}^{(t)}, \mathbf{v}^{(t)})
\end{equation}
With $\mathbf{W}^{(t+1)}$, $\mathbf{U}^{(t+1)}$ and $\mathbf{Z}^{(t+1)}$ fixed, the objective function $\mathcal{J}(\mathbf{U}^{(t+1)},\mathbf{Z}^{(t+1)},\mathbf{W}^{(t+1)}, \mathbf{v})$ degenerates to the self-paced function \cite{kumar2010self,jiang2014easy}. Resorting to the Lemma \ref{lem2}, we have:
\begin{equation}
    \begin{split}
       \sum_{j=1}^m& \mathbf{v}^{(t+1)}_j\left(\sum_{k=1}^K\mathbf{W}_{jk}\|\mathbf{x}_j^t-\mathbf{U}_kZ_j^k\|_F^2-\lambda\right)\\
       \leq&\mathbf{v}^{(t)}_j\left(\sum_{k=1}^K\mathbf{W}_{jk}\|\mathbf{x}_j^t-\mathbf{U}_k\mathbf{Z}_j^k\|_F^2-\lambda\right)
    \end{split}
\end{equation} 
Thus, $\mathbf{v}^{(t+1)}$ is the the global optimum for $\mathcal{J}(\mathbf{U}^{(t+1)},\mathbf{Z}^{(t+1)},\mathbf{W}^{(t+1)}, \mathbf{v})$, which satisfies the following inequality:
\begin{equation}\label{eqt3}
    \mathcal{J}(\mathbf{U}^{(t+1)},\mathbf{Z}^{(t+1)},\mathbf{W}^{(t+1)}, \mathbf{v}^{(t+1)})\leq \mathcal{J}(\mathbf{U}^{(t+1)},\mathbf{Z}^{(t+1)},\mathbf{W}^{(t+1)}, \mathbf{v}^{(t)})
\end{equation}
In summary, Eqs.(\ref{eqt1}), (\ref{eqt2}) and (\ref{eqt3}) indicate that the objective value decreases in every iteration. Further, since the objective function itself is the sum of finite elements, thus it is bounded from below. Consequently, according to \cite{tseng2001convergence}, the Algorithm \ref{alg1} will be convergent, Theorem \ref{the1} has been proved.
\section{Experiments\label{s4}}
In this section, we evaluate the performance of PAS over the closed-set UDA and the PDA settings on the widely-used benchmark datasets. To evaluate the effectiveness of PAS, we compare it with several state-of-the-art UDA approaches and also provide an ablation study to demonstrate the effectiveness of the progressive adaptation in PAS. Moreover, we investigate the effect of the hyper-parameter on performance.
\subsection{PAS on Closed-Set UDA }
\subsubsection{Dataset}
We first verify the effectiveness of PAS on the setting of the closed-set UDA over two datasets including \emph{Office-Home}  and \emph{Image-Clef}, which are both the commonly-used benchmark datasets for the closed-set UDA and widely adopted in the most existing works such as \cite{pan2010domain,ganin2015unsupervised,long2017deep,long2015learning,venkateswara2017deep}. The statistics of the so-involved datasets are given as follows.\\
\noindent
\textbf{Office-Home} \cite{venkateswara2017deep} contains 15588 images of 65 categories from 4 domains: Artistic images (Ar), Clipart images (Cl), Product images (Pr), and Real-world images (Rw).\\
\noindent
\textbf{Image-CLEF} \cite{long2017deep} derives from Image-CLEF 2014 domain adaptation challenge, and is organized by selecting 12 object categories shared in the three famous real-world datasets, ImageNet ILSVRC 2012 (I), Pascal VOC 2012 (P), Caltech-256 (C). It includes 50 images in each category and totally 600 images for each domain.
\subsubsection{Experimental Setup}
In order to verify the effectiveness of PAS on closed-set UDA, we compare PAS respectively with several traditional and deep-learning-based adaptation  state-of-the-arts: 1 Nearest Neighbor (1NN), Transfer Component Analysis (TCA) \cite{pan2010domain}, Transfer Joint Matching (TJM) \cite{long2013adaptation}, Balanced Domain Adaptation (BDA) \cite{wang2017balanced}, Correlation Alignment (CORAL) \cite{sun2016return}, Geodesic Flow Kernel (GKF) \cite{gong2012geodesic}, Subspace Alignment (SA) \cite{fernando2013unsupervised}, ResNet50 \cite{he2016deep}, Deep Adaptation Network (DAN) \cite{long2015learning}, Domain Adversarial Neural Networks (DANN) \cite{ganin2015unsupervised}, Joint Adaptation Networks (JAN) \cite{long2017deep} and Conditional Adversarial Networks (CDAN) \cite{long2018conditional}. Specifically, we use PAS(c) to denote the results of the initialized subspaces. \\
The results of the deep-learning-based approaches (e.g., DAN, DANN, JAN and CDAN) are obtained directly from the existing works \cite{ganin2015unsupervised,long2018conditional,long2017deep,long2015learning}. For fair comparison, we use the 2048-dimensional deep feature (extracted using ResNet50 pre-trained on ImageNet) for both PAS and other shallow UDA approaches. The optimal parameters of all compared methods are set according to their original papers. As for PAS, we empirically set the dimension of subspace $D=1$. To evaluate the performance, we follow the widely used \textbf{accuracy} as a measure \cite{long2013adaptation,cao2018partial,wang2017balanced}.

\begin{table*}[]
\tiny

    \centering
    \begin{tabular}{| c | c |c |c |c | c | c | c | c | c | c | c | c | c | c | c|}
\hline
       $~~$ & ResNet & 1NN  & TCA  & TJM  & BDA  & CORAL & GFK  & SA   & DAN  & DANN & JAN  & CDAN  & PAS (c)    & PAS           \\  \hline \hline
        Ar$\rightarrow$ Cl & 34.9   & 45.3 & 38.3 & 38.1 & 38.9 & 42.2  & 38.9 & 43.6 & 43.6 & 45.6 & 45.9 & 46.6 & {\ul 51.8} & \textbf{52.2}\\ \hline
        Ar$\rightarrow$ Pr & 50.0   & 57.0 & 58.7 & 58.4 & 57.1 & 59.1  & 57.1 & 63.3 & 57.0 & 59.3 & 61.2 & 65.9          & {\ul 68.4} & \textbf{72.9} \\ \hline
       Ar$\rightarrow$ Rw & 58.0   & 45.7 & 61.7 & 62.0 & 60.1 & 64.9  & 60.1 & 68.0 & 67.9 & 70.1 & 68.9 & {\ul 73.4}    & 73.1       & \textbf{76.9} \\ \hline
       Cl$\rightarrow$ Ar & 37.4   & 57.0 & 39.3 & 38.4 & 38.7 & 46.4  & 38.7 & 47.7 & 45.8 & 47.0 & 50.4 & {\ul 55.7}    & 55.3       & \textbf{58.4} \\ \hline
       Cl$\rightarrow$ Pr & 41.9   & 58.7 & 52.4 & 52.9 & 53.1 & 56.3  & 53.1 & 60.7 & 56.5 & 58.5 & 59.7 & 62.7          & {\ul 63.3} & \textbf{68.1} \\ \hline
       Cl$\rightarrow$ Rw & 46.2   & 48.1 & 56.0 & 55.5 & 55.5 & 58.3  & 55.5 & 61.9 & 60.4 & 60.9 & 61.0 & 64.2          & {\ul 66.5} & \textbf{69.7} \\ \hline
       Pr$\rightarrow$ Ar & 38.5   & 42.9 & 42.6 & 41.5 & 42.2 & 45.4  & 42.2 & 48.2 & 44.0 & 46.1 & 45.8 & 51.8          & {\ul 55.5} & \textbf{58.3} \\ \hline
       Pr$\rightarrow$ Cl & 31.2   & 42.9 & 37.5 & 37.8 & 37.6 & 41.2  & 37.6 & 41.5 & 43.6 & 43.7 & 43.4 & \textbf{49.1} & 46.9       & {\ul 47.4} \\ \hline
       Pr$\rightarrow$ Rw & 60.4   & 68.9 & 64.1 & 65.0 & 64.6 & 68.5  & 64.6 & 70.0 & 67.7 & 68.5 & 70.3 & {\ul 74.5}    & 74.0       & \textbf{76.6} \\ \hline
       Rw$\rightarrow$ Ar & 53.9   & 60.8 & 52.6 & 53.0 & 53.8 & 60.1  & 53.7 & 59.4 & 63.1 & 63.2 & 63.9 & \textbf{68.2} & 65.5       & {\ul 67.1}   \\ \hline
       Rw$\rightarrow$ Cl & 41.2   & 48.3 & 41.7 & 42.0 & 42.3 & 48.2  & 42.3 & 47.4 & 51.5 & 51.8 & 52.4 & \textbf{56.9} & 52.9       & {\ul 53.5}          \\ \hline
       Rw$\rightarrow$ Pr & 59.9   & 74.7 & 70.5 & 71.4 & 70.6 & 73.1  & 70.6 & 74.6 & 74.3 & 76.8 & 76.8 & \textbf{80.7} & 76.9       & {\ul 77.6}          \\ \hline
       AVE & 46.1 & 56.4 & 51.3 & 51.3 & 51.2 & 55.3 & 51.2 & 57.2 & 56.3 & 57.6 & 58.3 & {\ul 62.8} & {\ul 62.8} & \textbf{64.9}\\

\hline
    \end{tabular}
    \caption{Accuracy on \emph{Office-Home} for unsupervised domain adaptation with closed-set}
    \label{table1}
\end{table*}

\begin{table*}[]
\tiny

\setlength{\parindent}{2em}
    \centering
    \begin{tabular}{| c| c |c |c |c | c | c | c | c | c | c | c | c | c | c | c |}
\hline
$~~$  & ResNet & 1NN  & TCA  & TJM  & BDA  & CORAL & GFK  & SA   & DAN  & DANN & JAN  & CDAN          & PAS (c)    & PAS           \\ \hline \hline
C$\rightarrow$I   & 78.0   & 83.5 & 89.3 & 90.0 & 90.8 & 83.0  & 86.3 & 88.2 & 86.3 & 87.0 & 89.5 & {\ul 91.3}          & 86.5 & \textbf{90.5} \\ \hline
C$\rightarrow$P   & 65.5   & 71.3 & 74.5 & 75.0 & 73.7 & 71.5  & 73.3 & 74.3 & 69.2 & 74.3 & 74.2 & 74.2          & {\ul 74.3} & \textbf{75.5} \\ \hline
I$\rightarrow$C   & 91.5   & 89.0 & 93.2 & 94.2 & 94.0 & 88.7  & 93.0 & 94.5 & 92.8 & 96.2 & 94.7 & \textbf{97.7}    & 93.0       & {\ul 95.1} \\ \hline
I$\rightarrow$P   & 74.8   & 74.8 & 77.5 & 76.2 & 75.3 & 73.7  & 75.5 & 76.8 & 74.5 & 75.0 & 76.8 & {\ul 77.7}    & {\ul 77.7}       & \textbf{78.3} \\ \hline
P$\rightarrow$C   & 91.2   & 76.2 & 83.7 & 85.3 & 83.5 & 72.0  & 82.3 & 93.5 & 89.8 & 91.5 & 91.7 & {\ul 94.3}          & 92.2 & \textbf{95.5} \\ \hline
P$\rightarrow$I   & 83.9   & 74.0 & 80.8 & 80.3 & 77.8 & 71.3  & 78.0 & 88.3 & 82.2 & 86.0 & 88.0 & {\ul 90.7}          & 89.7 & \textbf{92.0} \\ \hline
AVE   & 80.7   & 78.1 & 83.2 & 83.5 & 85.5 & 76.7  & 81.4 & 85.9 & 82.5 & 85.0 & 85.8 & {\ul 87.7}          & 85.6 & \textbf{87.8} \\ \hline
    \end{tabular}
    \caption{Accuracy on \emph{Image-Clef} for unsupervised domain adaptation with closed-set}
    \label{table2}
\end{table*}

\subsubsection{Experimental Results}
The classification results of 12 tasks on \emph{Office-Home} dataset and 6 tasks on \emph{Image-Clef} dataset are shown in Tables \ref{table1} and \ref{table2}, respectively. From Tables \ref{table1} and \ref{table2}, we can see that PAS achieves the highest average accuracy on both two datasets. Specifically, PAS achieves the best average accuracy of 64.9\% and achieves the best or the second-best performance in all 12 tasks on \emph{Office-Home} dataset. The proposed PAS approach performs the best or the second-best in all 6 tasks and ranks the first place with the average accuracy of 87.8\% on the \emph{Image-Clef} dataset. Moreover, since these results are obtainedacross many image datasets used here, the results also demonstrate that the proposed PAS approach is capable of knowledge transferring in domain adaptation.

\subsection{PAS on PDA}
\subsubsection{Dataset}
For evaluating the performance of PAS on the PDA setting, we further conduct experiments on the commonly-used PDA benchmarking dataset \emph{Office-31} \cite{cao2018partial,cao2018partialCVPR,zhang2018importance}, whose detail is given as follows:\\
\noindent
\textbf{Office-31} (\cite{saenko2010adapting} contains 4652 images with 31 categories in three visual domains Amazon(A), DSLR(D), Webcam(W). For the setting of PDA, we follow the same splits used in recent PDA studies \cite{cao2018partial,cao2018partialCVPR}. Specifically, each source domain contains 31 categories and each target domain contains 10 categories (backpack, bike, calculator, head-phone, computer-keyboard, laptop-101, computer-monitor, computer-mouse, coffee-mug, and video-projector), which is shared with Caltech-256.
\subsubsection{Experimental Setup}
To evaluate the performance of PAS on the PDA setting, we compare PAS with several PDA state-of-the-arts: Selective Adversarial Network (SAN) \cite{cao2018partialCVPR}, Importance Weighted Adversarial Network (IWAN) \cite{zhang2018importance}, Partial Adversarial Domain Adaptation (PADA) \cite{cao2018partial}. Two Weighted Inconsistency-reduced Networks (TWINs) \cite{matsuura2018twins} and Example Transfer Network (ETN) \cite{cao2019learning}. To the best of our knowledge, so far, there is no shallow approaches proposed for PDA setting. In PDA, we also use the 2048-dimensional deep features (extracted using ResNet50 pre-trained on Image-Net) for PAS. All results of the compared methods are obtained from the existing works \cite{cao2019learning,matsuura2018twins}. For PAS, we empirically set the dimension of subspace $D=10$ for PDA setting. 
\subsubsection{Experimental Results}
The classification results of the 6 PDA tasks on \emph{Office-31} dataset are given in Table \ref{table3}. Specifically, our approach PAS outperforms all state-of-the-arts with an average accuracy of 97.2\% by using a simple subspace learning method, while the second-highest average accuracy (96.7\%) ETN is achieved by the artificially designed instance re-weighting strategy and the complex deep feature learning scheme. In addition, almost all closed-set UDA approaches (e.g., TJM, DAN, DANN) work even worse than the baseline methods without DA (i.e., 1NN), since the risk of mode collapse is highly increased in the PDA setting. The results demonstrate the robustness of PAS for adaptation under the large shift (some categories do not exist in the target domain), which also illustrates the importance of mitigating the mode collapse while adaptation.

\begin{table*}[]
\tiny

\setlength{\parindent}{2em}
    \centering
    \begin{tabular}{| c | c |c |c |c | c | c | c | c | c | c | c | c | c | c |c |}
\hline
$~~$& ResNet & 1NN  & TJM  & GFK  & SA   & DAN  & DANN & PADA          & TWINs & IWAN       & SAN           & ETN           & PAS(c) & PAS           \\
\hline\hline
A$\rightarrow$W   & 54.5   & 82.0 & 81.2 & 74.6 & 73.2 & 46.4 & 41.4 & 86.5          & 86.0  & 89.2       & 93.9          & 94.5          & 75.9   & \textbf{97.0} \\ \hline
D$\rightarrow$W   & 94.6   & 75.6 & 74.8 & 96.6 & 87.1 & 53.6 & 46.8 & 99.3          & 99.3  & 99.3       & 99.3          & 100           & 98.3   & {\ul 99.3}    \\  \hline
W$\rightarrow$D & 94.3 & 71.2 & 93.2 & 99.4 & 96.8 & 58.6 & 38.9 & \textbf{100} & \textbf{100} & 99.4 & \textbf{100} & \textbf{100} & \textbf{100} & \textbf{100}  \\ \hline
A$\rightarrow$D   & 65.6   & 79.2 & 74.4 & 82.8 & 82.2 & 42.7 & 41.4 & 82.2          & 86.8  & 90.5       & 82.2          & {\ul 95.0}    & 85.4   & \textbf{98.4} \\ \hline
D$\rightarrow$A   & 73.2   & 82.4 & 83.2 & 82.7 & 79.1 & 65.7 & 41.3 & 92.7          & 94.7  & {\ul 95.6} & 92.7          & \textbf{96.2} & 85.5   & 94.6          \\ \hline
W$\rightarrow$A   & 71.7   & 76.4 & 71.6 & 83.8 & 87.0 & 65.3 & 44.7 & \textbf{95.4} & 94.5  & 94.3       & \textbf{95.4} & 94.6          & 87.9   & 94.4          \\ \hline
AVE   & 75.6   & 81.1 & 79.7 & 86.6 & 84.2 & 55.4 & 42.4 & 92.7          & 93.6  & 94.7       & 92.7          & {\ul 96.7}    & 88.8   & \textbf{97.2} \\ \hline
    \end{tabular}
    \caption{Accuracy on \emph{Office-31} for unsupervised domain adaptation from 31 classes to 10 classes}
    \label{table3}
\end{table*}
In summary, on our experimental datasets, the proposed PAS approach can outperform both deep and shallow UDA approaches on both closed-set UDA and PDA settings. Compared to the elaborately designed UDA approaches, PAS only involves one simple subspace learning technique(i.e., PCA) and one simple sample anchoring scheme (i.e., $l_1$-norm), which further implies the effectiveness of progressive adaptation. 

\subsection{Analysis of Progressive Adaptation} \label{ss1}
\begin{figure}[t]
    \centering
     \includegraphics[width=1\textwidth]{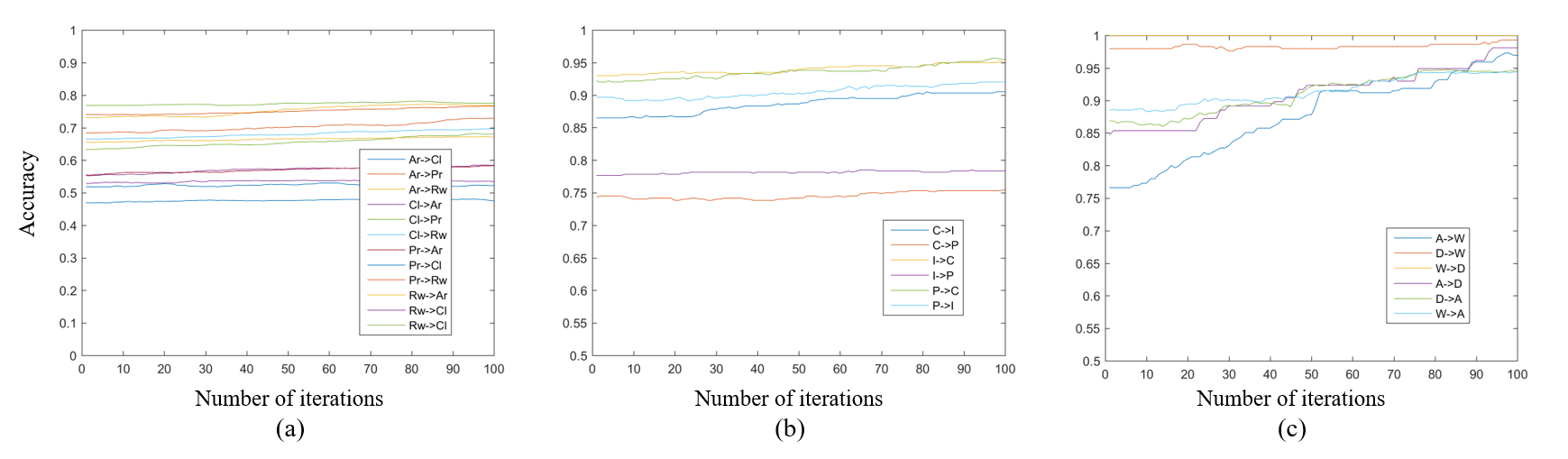}
     \caption{The accuracy of the pseudo labels with progressively refining adaptation of subspaces on (a) \emph{Office-Home},(b)  \emph{Image-Clef} and (c) \emph{Office-31} datasets.}
    \label{figure2}
\end{figure}

In order to demonstrate the effectiveness of the proposed progressive adaptation mechanism, we perform an ablation study that evaluates the subspace learning without refinement (i.e., initial subspaces learned from source domain) which is marked as PAS (c) in Tables \ref{table1}, \ref{table2} and \ref{table3}. It can be observed that progressive adaptation mechanism outperforms the PAS(c). The results illustrate that the proposed progressive adaptation mechanism can effectively convince the pseudo-labels and thereby improve the performance of adaptation.  \\
Moreover, in Figure \ref{figure2}, we also examine the classification accuracy of the pseudo labels in each  progressive iteration on \emph{Office-Home}, \emph{Image-Clef} and \emph{Office-31} datasets to further understand the mechanism of progressive adaptation. As shown in Figure \ref{figure2}, we observe that the accuracy or the quality of the pseudo labels is gradually improved with the progressive target samples anchoring and subspace refining, confirming that progressive adaptation can desirably mitigate the risk of mode collapse in adaptation. In addition, the result in Figure \ref{figure2} also clearly indicates that \emph{anchoring target samples to refine the shared subspace is quite important}, since the support for the initial subspace is defined only over the source domain. 

\subsection{Analysis of Hyper-parameter}
\begin{figure}[h]
    \centering
    \setlength{\abovecaptionskip}{0pt}
    \setlength{\belowcaptionskip}{0pt}
     \includegraphics[width=1\textwidth]{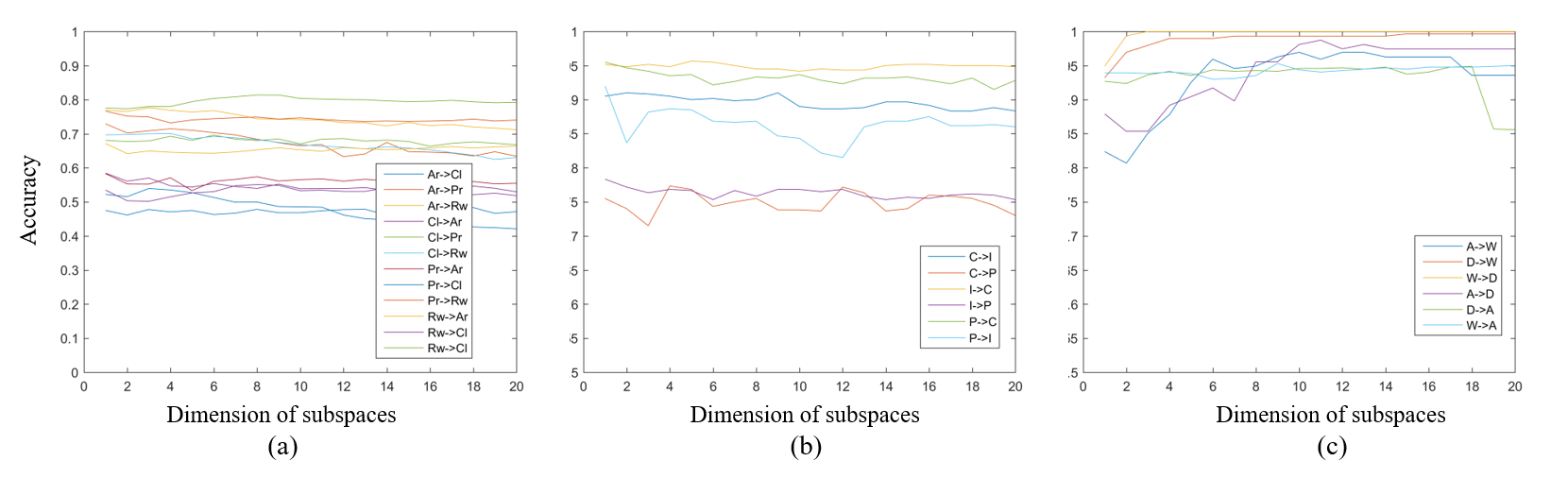}
     \caption{The accuracy of the pseudo labels with progressively refining adaptation of subspaces on (a) \emph{Office-Home},(b) \emph{Image-Clef} and (c) \emph{Office-31} datasets.}
    \label{figure3}
\end{figure}

The proposed PAS method contains one hyper-parameter( i.e., dimension of subspace $D$). To investigate the effect of the subspace dimension $D$ on performance, we conduct experiments on \emph{office-Home}, \emph{Image-Clef} and \emph{Office-31} datasets and report the accuracy of PAS by varying the dimension of subspaces from 1 to 20 with the increment. As we can see  in Figure \ref{figure3} (a - c), the Dimension $D$ of the subspace can perform robustly and insensitively on both the closed-set UDA and PDA tasks on a wide range of parameter values. Specifically, the best choice of $D$ on the closed-set UDA setting is $D\in[1,20]$ and on the PDA setting is $D\in[5,15]$.

\subsection{Analysis of the Sample Anchoring} \label{ss2}
 To illustrate the sample anchoring scheme in PAS, we analyze the Accuracy (Acc) and the Average Density Ratio (ADR) of the top and bottom 5\% samples according to the squared distance in Eq.(\ref{eq4}) on \emph{Office-Home}, \emph{Image-Clef} and \emph{Office-31} datasets as shown in Tables \ref{table4}, \ref{table5} and \ref{table6}, respectively.\\
Specifically, the density ratio $w(x)=\frac{p (x)}{q (x)}$ is estimated by minimizing the  Kullback-Leibler Divergence from  $p(x)$ to $w(x)q(x)$, which is also known as KLIEP \cite{tsuboi2009direct}. Accordingly, we adopt the toolbox shared by Sugiyama\footnote{\href{http://www.ms.k.u-tokyo.ac.jp/software/KLIEP/KLIEP.zip }{http://www.ms.k.u-tokyo.ac.jp/software/KLIEP/KLIEP.zip}.}  to solve $w(x)$ for all target samples. Then, Average Density Ratio of the top or bottom 5\% samples can be formulated as:
\begin{equation}
    \label{eq16}
          ADR=\frac{1}{na}\sum_{a=1}^{na}w(x_a)
\end{equation} 
where $na$ is the number of anchored bottom or top 5\% target samples. According to its definition, a large ADR indicates a high probability under the source distribution, but a low probability under the target domain \cite{tsuboi2009direct}. \\
In Tables \ref{table4}, \ref{table5} and \ref{table6}, the anchored top 5\% target samples with small squared distance tend to large ADR, while the anchored bottom 5\% targets tend to opposite results, illustrating that the anchored samples with small squared distance have high probability under the source distribution and thereby tend to low KL divergence with source domain, not vice verse. These results illustrate that the squared distance can effectively approximate the domain discrepancy to some extent. Moreover, such results also reveal that the target samples with small squared distance tend to have a large Acc, which is consistent with the theoretical result in recent study \cite{ben2010theory}. To this end, the proposed sample anchoring scheme can effectively localize the target samples with reliable pseudo labels and thereby effectively refine the shared subspaces.
\begin{table}[]
\tiny
\centering
\begin{tabular}{|c|c|c||c|c|}
\hline
\multirow{2}{*}{} & \multicolumn{2}{c||}{Top 5\%} & \multicolumn{2}{c|}{Bottom5\%} \\ \cline{2-5}
                  & Acc         & ADR           & Acc           & ADR            \\ \hline\hline
Ar$\rightarrow$Cl             & 91.78        & 15.8462       & 25.11         & 0.0001         \\ \hline
Ar$\rightarrow$Pr             & 97.30        & 12.4667       & 30.63         & 0.0002         \\ \hline
Ar$\rightarrow$Rw             & 99.54        & 10.2736       & 41.28         & 0.0049         \\ \hline
Cl$\rightarrow$Ar             & 95.08        & 15.7139       & 14.75         & 0.0000         \\ \hline
Cl$\rightarrow$Pr             & 97.18        & 9.4412        & 19.36         & 0.0003         \\ \hline
Cl$\rightarrow$Rw             & 97.25        & 8.0402        & 27.98         & 0.0002         \\ \hline
Pr$\rightarrow$Ar             & 96.72        & 14.9313       & 19.67         & 0.0002         \\ \hline
Pr$\rightarrow$Cl             & 97.98        & 13.0356       & 19.18         & 0.0014         \\ \hline
Pr$\rightarrow$Rw             & 99.54        & 6.1650        & 28.90         & 0.0017         \\ \hline
Rw$\rightarrow$Ar             & 99.18        & 7.9605        & 30.33         & 0.0005         \\ \hline
Rw$\rightarrow$Cl             & 78.54        & 11.6883       & 14.16         & 0.0010         \\ \hline
Rw$\rightarrow$Pr             & 96.85        & 5.1629        & 26.13         & 0.0085         \\ \hline
AVE               & 95.38        & 10.8938       & 24.83         & 0.0016         \\ \hline
\end{tabular}
    \caption{Accuracy (Acc) and Average density ratio (ADR) of top and bottom 5\% samples according to squared distance on \emph{Office-Home} dataset.}
    \label{table4}
\end{table}
\begin{table}[]
\tiny
\centering
\begin{tabular}{|c|c|c||c|c|}
\hline
\multirow{2}{*}{} & \multicolumn{2}{c||}{Top 5\%} & \multicolumn{2}{c|}{Bottom5\%} \\ \cline{2-5}
                  & Acc         & ADR           & Acc           & ADR            \\ \hline\hline
C$\rightarrow$I               & 100         & 4.6189        & 73.33         & 0.0069         \\ \hline
C$\rightarrow$P               & 100         & 17.4481       & 70.00         & 0.0011         \\ \hline
I$\rightarrow$C               & 100         & 3.5151        & 80.00         & 0.0001         \\ \hline
I$\rightarrow$P               & 100         & 13.4076       & 70.00         & 0.0061         \\ \hline
P$\rightarrow$C               & 96.69       & 13.3154       & 60.00         & 0.0020         \\ \hline
P$\rightarrow$I               & 100         & 15.7842       & 93.33         & 0.0006         \\ \hline
AVE               & 99.44       & 11.3482       & 74.44         & 0.0028         \\ \hline
\end{tabular}
    \caption{Accuracy (Acc) and Average density ratio (ADR) of top and bottom 5\% samples according to squared distance on \emph{Image-Clef} dataset.}
    \label{table5}
\end{table}
\begin{table}[]
\tiny
\centering
\begin{tabular}{|c|c|c||c|c|}
\hline
\multirow{2}{*}{} & \multicolumn{2}{c||}{Top 5\%} & \multicolumn{2}{c|}{Bottom5\%} \\ \cline{2-5} 
                  & Acc         & ADR           & Acc           & ADR            \\ \hline\hline
A$\rightarrow$W               & 87.21       & 15.5812       & 23.29         & 0.0004         \\ \hline
D$\rightarrow$W               & 96.40       & 12.9463       & 30.18         & 0.0005         \\ \hline
W$\rightarrow$D               & 93.45       & 15.9949       & 10.66         & 0.0000         \\ \hline
A$\rightarrow$D               & 95.95       & 9.6682        & 15.32         & 0.0013         \\ \hline
D$\rightarrow$A               & 91.80       & 15.6139       & 16.39         & 0.0002         \\ \hline
W$\rightarrow$A               & 94.06       & 12.7443       & 13.70         & 0.0015         \\ \hline
AVE               & 93.14       & 13.7581       & 18.26         & 0.0007         \\ \hline
\end{tabular}
    \caption{Accuracy (Acc) and Average density ratio (ADR) of top and bottom 5\% samples according to squared distance on \emph{Office-31} dataset.}
    \label{table6}
\end{table}

\section{Conclusion\label{s5}}
In this paper, we aim to mitigate the risk of mode collapse intuitively based on the progressive adaptation. Specifically, we proposed an effective UDA method named Progressive Adaptation of Subspaces (PAS). The experimental results on several benchmark datasets demonstrate the effectiveness of the proposed PAS on both UDA with closed set and PDA settings. These results confirm that progressive adaptation scheme is applicable to mitigate mode collapse in adaptation. In the end, this study also indicates that domain alignment may not be the only method for solving the UDA problem, which will be further studied in our future work.


\section*{ACKNOWLEDGMENTS}

This work is supported in part by Key Program of NSFC under Grant No. 61732006.


\bibliography{mybibfile}

\end{document}